\documentclass{IOS-Book-Article}

\usepackage{mathptmx}
\usepackage{soul}\setuldepth{article}
\usepackage{graphicx}
\usepackage{natbib}
\usepackage{comment}
\def\hb{\hbox to 11.5 cm{}}

\begin{document}

\pagestyle{plain}

\begin{frontmatter}              

\title{Symbiotic Child Emotional Support with Social Robots and Temporal Knowledge Graphs}

\markboth{}{May 2022\hb}
\author[A]{\fnms{Isabella Saccardi}}\footnote{The authors contribute equally to this paper.},
\author[A]{\fnms{Duygu Sezen Islakoglu}}\footnotemark[\value{footnote}],
\author[A]{\fnms{Anouk Neerincx}},
\author[A]{\fnms{Federica Lucia Vinella}}
\runningauthor{Saccardi, Islakoglu, et al.}
\address[A]{Utrecht University, Department of Information and Computing Sciences, Utrecht, The Netherlands}


\begin{abstract}
In current youth-care programs, children with needs (mental health, family issues, learning disabilities, and autism) receive support from youth and family experts as one-to-one assistance at schools or hospitals. Occasionally, social robots have featured in such settings as support roles in a one-to-one interaction with the child. In this paper, we suggest the development of a symbiotic framework for real-time Emotional Support (ES) with social robots Knowledge Graphs (KG). By augmenting a domain-specific corpus from the literature on ES for children (between the age of 8 and 12) and providing scenario-driven context including the history of events, we suggest developing an experimental knowledge-aware ES framework. The framework both guides the social robot in providing ES statements to the child and assists the expert in tracking and interpreting the child's emotional state and related events over time.  

\end{abstract}

\end{frontmatter}
\markboth{June 2021\hb}{June 2021\hb}

\section{Introduction}
Social robots are increasingly popular as Emotional Support (ES) actors in youth-care programs for children (e.g., in hospitals \citep{jeong2015social, moerman2021using, meghdari2016conceptual}, schools \citep{kirstein2016social, belpaeme2018social}, and youth-care centers \citep{saldien2010expressing, huijnen2016mapping}). 
Yet, it is still unclear which role the social robot should entail and what a fulfilling child-robot relationship looks like in this domain. Experts in youth care and families have expressed the need to increase knowledge regarding the risks and opportunities of employing social robots in youth healthcare practices. Consequentially, scientific empirical studies are beginning to advance robot-supported ES programs for youth to establish regulated and safe practices \citep{neerincx2021social, moerman2019social}.

\citet{neerincx2021social} for example, evaluated the robot-child interaction as a means to break the ice, run standardized tests, and play educational games in settings such as hospitals and schools.
Several other studies have proposed methods to establish empathy in the child-robot interaction to aid social presence, engagement, and social support (e.g., \citep{leite2014empathic, ligthart2018reducing, belpaeme2012multimodal}). Jointly, the general public is shifting toward the notion of assistive social robots as capable of a deeper, more intimate understanding when communicating with their human counterparts \citep{gamborino2019mood}.

Methodologically, researchers have deployed state-of-the-art machine learning techniques such as Reinforcement Learning to form mood estimation as a social profile predictor in social robots \citep{gamborino2019mood}. Withal, these techniques often lack explainability and interpretability \citep{puiutta2020explainable} due to implicit policies and hidden processes the robot relies upon to make decisions. The outcome is a black-box phenomenon where humans do not know how or why the AI solved a given problem the way it did \citep{adadi2018peeking}. We present a way to deal with the specific problem of AI as a black box within the robot-child ES domain by proposing the development of a semantic representation of child events. We intend to guide and adapt the robot-child interaction while informing the expert of the child's emotional history. Our framework relies on the domain-specific Knowledge Graph (KG) construct to map events into triples (Entity-Relation-Entity). These triples would assist the robot in tailoring the ES statements to the child. They would also inform the expert by monitoring the temporal evolution of the child's emotional state.

Knowledge Graphs (KG) -- also known as semantic networks -- represent networks of real-world entities (i.e., objects, events, situations, or concepts) and illustrate the relationship between them \citep{Ji_2022}. This paper focuses on generic KGs based on triples (subject, relation, object) and Temporal KGs (TKG) where each triple is associated with a timestamp or time interval and largely models evolving social networks. KGs are already widely used in applications such as personal assistants, question answering, and recommendation systems \citep{wang2017knowledge}. In the human-robot paradigm, KGs have been studied for human-robot collaboration \citep{ding2019robotic, zhang2021brain}, human-robot knowledge transfer \citep{shukla2015unified}, and situational human-robot interaction \citep{zhang2021patterns}. However, robot-aided child ES supported by KGs and TKGs is altogether a novel application.

\textbf{The contribution of the proposed framework is two-fold. One is to supply adaptive real-time ES to the child through the social robot; the other is to give contextualized knowledge to the expert}. Firstly, we suggest developing an abstract KG built on the literature on child ES that we will construct from triples. Next, we plan to generate an annotated ES corpus from crowdsourcing annotation (\citep{smith2014development, dennis2013towards, liu2021towards}) motivated by the scenarios built from the abstract KG. The corpus would guide the prediction of adaptive ES statements. \textbf{Our framework is an example of a symbiotic relationship between robots and humans}. Here, the child and the expert keep the social robot up-to-date while the social robot provides emotional support to the child and assists the expert. Thanks to their interpretable nature, KGs are ideal for information visualization and explanation aiding with the development of human-centered robot-child ES. 

The rest of the paper is as follows. Section \ref{sec:experimental} presents the five phases of the experimental design, the plan, and the procedure. Section \ref{lim&chall} discusses the limitations and challenges of the proposed framework. Section \ref{sec:conc} concludes the work.

\section{Experimental Design, Plan and Procedure} \label{sec:experimental}
In this paper, we propose a research pipeline divided into five phases namely 1. instantiation of KGs, 2. creation of ES corpus, 3. knowledge update, 4. support, and 5. evaluation. While all five phases are designed to follow each other linearly, Phase 3 and Phase 4 will occur in a loop and reinforce each other. Each participant in the current research will be appropriately informed about the scope of the research and the data usage.

\subsection{Phase 1: Instantiation of KGs}
\label{phase1} 

In this preliminary phase, we propose focusing on the instantiation of two knowledge graphs. One deals with the theoretical grounding of stressors and emotional states in children and is an abstract and generalized KG. The second is a TKG offering a temporal overview of a child's emotional states and related situations.

 \textit{Abstract Knowledge Graph.} Firstly, we suggest running an extensive systematic literature analysis on child ES research to gather theories on child stressors coupled with qualitative coding, thematic analysis, and ground theory. The result will be an abstract Knowledge Graph (KG) that will consist of Entity-Relation-Entity triples on child stressors and emotional states known within the psychological domain (e.g., child (\textit{Entity}) refusing to go (negative emotion/\textit{Relation}) to school (\textit{Entity})). We propose using existing techniques to build multiple scenarios of child stressors from the abstract KG. The abstract KG is herein the ground-truth and constant featuring in all child ES programs.
Once assessed and approved by a set of invited experts (including experts in pediatrics, education, and pedagogy), Phase 2 of the pipeline will utilize this validated abstract KG to form scenarios for creating a domain-specific ES corpus.

\textit{Child Temporal Knowledge Graph.} The second type of graph will be a Temporal KG (TKG) instantiated at the time of the first intake meeting with the child and the expert. The expert will ensure that the entities of this graph will have corresponding entities in the abstract KG. In other words, the abstract KG will serve as a base for the child TKG. The child TKG instance will update during the follow-up sessions with the expert and the social robot (Phase 3). It will comprise the history of social interactions (e.g., family and friends) and timestamps/time intervals.

\subsection{Phase 2: Creation of ES corpus}
\label{phase2} 

We suggest developing a corpus of ES statements by crowdsourcing human annotation. This phase is divided into sub-phases namely Scenarios, Corpus creation, Corpus validation, Corpus classification, and Corpus augmentation.

    \paragraph{Scenarios.} From the abstract KG formed in Phase 1, we propose focusing on a specific set of triples characterized by negative or distressing situations. The selected triples will be used as use cases for a set of hypothetical scenarios where the child would typically receive ES.  
    
    \paragraph{Corpus creation.} With the given collection of scenarios, we propose to recruit crowd workers to generate appropriate ES statements and assign an ES category to each triple. An example of a question for this crowdsourcing task would be \textit{"Write down three possible sentences that you would say to a child who is experiencing this situation."}. The resulting support statements will form an initial unverified corpus of ES statements.
    
    \paragraph{Corpus validation.} Afterward, the corpus will be validated by an independent set of crowd workers. This sub-phase is needed to ensure that the selected statements are appropriate for the ES program and the child. Crowd workers will rate on a five-point Likert scale the appropriateness of the content for children of 8 to 12 years old. Statements with low rating or high standard deviation (e.g., \textit{mean$<$ given threshold}, \textit{sd$>$given threshold}) will be excluded from the final corpus. The resulting body of statements will form a new, validated ES corpus.
   
    \paragraph{Corpus classification.}  In a follow-up crowdsourcing task, each crowd worker will be asked to categorize a subset of statements (approx. 20) from the validated corpus with one of the ten chosen classes from the literature (\citep{smith2014development}\footnote{The category of Deserving was based on reminding the distressed person of what they deserve (e.g., some time off, some time with their friends). 
    This category was considered too complex for children and therefore excluded.}, \citet{liu2021towards}\footnote{\citet{liu2021towards} categories - named strategies in their article - partially overlap with the ones of \citet{smith2014development}, except for the Information strategy, which aims to provide information about the situation. This was excluded due to limitations of the system, as providing real-time complete information will likely not be possible.}, \citet{dennis2013towards} \footnote{The category of Reassurance was added, as deemed suitable for children.}, see Table \ref{tab:EScategories}). The intended result will be an annotated and validated ES corpus where each triple is associated with one scenario which is categorized with one class. To achieve this, we plan to calculate  intra-rater agreement, and propose using a free-marginal kappa \citep{randolph2005free}. Statements categorized with a marginal \textit{kappa $<$ given threshold} will be excluded from the corpus as too ambiguous.
   
    \paragraph{Corpus augmentation.} We expect the final annotated corpus to be of limited size (not exceeding 100 statements) and sub-divided into ten ES categories. To ensure a higher variability of the sentences, we propose to deploy a pre-trained language model to augment the corpus by paraphrasing. Then, the expert will review the final corpus: any sentence deemed inappropriate will be excluded. After the validation of the new instances, the resulting augmented ES corpus will be used in the next phase of this research and will be continuously updated with exceptions encountered by the robot during the interaction with the child (see Phase 4).




\subsection{Phase 3: Knowledge update}
\label{phase3} 

In this phase, the child TKG updates with the child-robot and expert-robot interactions. The robot-child interaction will create new triples with time information as the child talks about past events and expresses emotions about events. If the time information is missing or is inaccurate, the robot will disambiguate by asking questions during the interaction. Thanks to its updating property, the child TKG will be robust and adaptable to changes.
Similarly, the expert will moderate the child TKG by applying changes to the graph whenever needed (e.g., by adding or fixing triples). Both kinds of child TKG updates (automated and manual) will support the notion of an \textbf{expert-in-the-loop approach}.


\subsection{Phase 4: Support}
We divide this phase into two types of support. One deals with the ES offered to the child. The other consists of the support that the expert receives from the framework in the form of context and knowledge. We explain these two kinds support in the following paragraphs.

 \paragraph{Providing adaptive support to the child.}
    This support deals with the ES sentences dispensed by the robot during the interaction with the child. The child's sentences will be converted to triples and saved in the child's TKG. Then, a similarity function \footnote{The similarity function will return a similarity score calculated by comparing two triple embeddings.} will decide whether this triple corresponds to an edge in the abstract KG.\footnote{Both abstract KG and child TKG will be heterogeneous as they will contain different types of entities and relations. However, we expect to encounter a limited number of entity types (parents, friends, teachers, etc.) and a higher number of relation types. The latter will increase the difficulty of calculating the similarity score between two triples.}   
    If the similarity function returns a sufficiently high similarity score (determined by a proposed confidence threshold), the model will use the corresponding ES category to produce a randomly picked ES statement matching that category. This way, the ES statement can always be justified by the abstract KG and, ultimately, by the literature. We suggest keeping a log of the ES statements used to avoid redundancy and dispense a semi-natural dialog between the robot and the child.


\paragraph{Providing support to the expert.}
	The framework makes the expert and the social robot dynamic partners that learn from each other.
	From the child's TKG, the expert will be able to gain several benefits.

 \textit{(a) Access contextual history and handle exceptions.}
	    The child TKG enables the extraction of statistics about the child's states over time. The expert could have access to a variety of information, for instance, the duration of a specific relation (e.g., feeling angry), the frequency of the events, the gaps between interactions, and the patterns in relational order (e.g., \textit{"Whenever a homework assigned, Child feels depressed."}). The expert can determine which outliers to discard according to an arbitrary threshold. Overall, statistical data will help experts catch and handle exceptions (i.e., outliers, missing information).
	    
\textit{(b) Predict patterns and assess the effectiveness of the ES program.}
	    Thanks to data-driven state-of-the-art TKG methods \citep{https://doi.org/10.48550/arxiv.2201.08236}, it is possible to predict the emotional state of the child within a given timeline. Since we plan to handle emotions as relations (e.g., feeling happy) in the child's TKG, a link prediction task could help the expert assess the probability of an emotional state for any given time. Specifically, the child's TKG can help compute the probability of a quadruple (Child, feeling happy, Child, Date) to manifest at some point in time. Using prediction methods such as Graph Neural Networks \citep{9046288}, the events that are close to a given date and  neighbor entities like parents will contribute to the prediction. The evolution of the emotional states will grant insights into the effects of the support program.

\textit{(c) Complete the knowledge about the child while preserving privacy.}
	    Another advantage of having the child TKG is the creation of anonymized scenarios when the expert decides to consult other authorities. Temporal walks from child entities can create chronological stories carrying rich contextual information \footnote{Information taken from the subgraph that covers relevant interactions with relevant entities over time.} while preserving privacy. Hence, an expert can utilize temporal walks to obtain better-tailored suggestions.

\subsection{Phase 5: Evaluation}
From the evaluation, we intend to assess the capacity of the proposed framework to function in real-life scenarios and within the domain herein presented. We also suggest evaluating ways that the experts and the children perceive the child-robot interaction with the given framework according to a set of criteria.
For evaluating the framework, we suggest utilizing both computational evaluative methods used for knowledge graphs (i.e. accuracy for classification, average precision for future link prediction) and qualitative methods established in the field of Human-Computer-Interaction (i.e.perceived anthropomorphism, animacy, likeability, perceived intelligence, and perceived safety of the robot \citep{bartneck2009measurement}). 

\label{phase5} 






\section{Limitations and challenges}
\label{lim&chall}

    \paragraph{Data sparsity.} Our proposed version of child TKG strongly relies on the robot-child interaction to generate triples. However, when the interaction is not common -- which can likely occur in real-life settings -- the TKG may not sufficiently capture the events that are thought relevant for the ES program. 
    For this reason, we believe that the expert can play a fundamental role in mitigating the lack of information (e.g., due to the cold start problem, outliers, data sparsity, etc.) since they can monitor the state of the child's KG and enrich it with complementary information from other sources.
    \paragraph{Accuracy.}  False negatives might occur when the ES framework does not identify input and does not provide an appropriate ES statement. False-positive might occur when the ES statement tailors a virtually nonexistent problem or the causal relationship is incorrect. For both types of prediction errors, the role of the expert will be crucial to correct the prediction and to complete the missing knowledge in graph. 
    
    \paragraph{Ambiguity.} Ambiguity can happen when the interaction is novel or when it is not clear how it can be translated into triple form. This issue can be prevented by programming the robot to ask questions when ambiguity is present in the child's TKG (e.g., triplets with ambiguous sentiment).
    
    \paragraph{Modality.} Emotion recognition encompasses other modalities such as gestural interaction, facial expressions, and body posture. These modalities can capture sentiment that children may find difficult to verbally express  \citep{neerincx2020social}. However, given the complexity of the domain, we suggest experimenting with the proposed language-based approach and in the future, introduce non-verbal modalities to complete the graphs.




\section{Conclusions} \label{sec:conc}
We present an experimental framework based on Knowledge Graphs providing Emotional Support (ES) to children aged 8 to 12 through child-robot interaction in schools and youth care centers. The research pipeline is divided into several phases, spanning from the instantiation of Knowledge Graphs to the generation and validation of an ES corpus. The end result is an adaptive framework for child-robot interaction that adapts to the child's emotional states over time. It also allows the expert to obtain contextualized knowledge to improve the ES program. Ours is a proposition for a mutualistic symbiotic relationship between the robot, the expert, and the child through semantic knowledge creation where each entity can mutually benefit from the interaction. 

\bibliography{main.bib}
\section{Appendix}

\begin{table}[h]
\begin{tabular}{p{3cm}p{5cm}p{3cm}}
\hline \textbf{Category}      & \textbf{Explanation}                                                                           & \textbf{Example}  \\ \hline
Appreciated (APP)      & This statement reminds the child of his / her importance.                                      & I’m glad you are here for..                     \\
Supported (SUP)        & This statement makes the child feel supported.                                                 & I am here for you.                              \\
Praise (PRS)           & This statement praises the child, focusing on his / her good qualities.                                                          & You are a good child.                           \\
Encouragement (ENC)    & This statement encourages the child to overcome the situation.                                 & You can do this.                                \\
Empathy (EMP)          & This statement legitimize the child's feeling, making him / her feel understood.               & I understand you are sad.                       \\
Practical Advice (PRA) &  This statement provides practical advices to deal with the situation.                          & Just do your best.                              \\
Emotional Advice (EMO) & This statement suggests to the child how to deal with his / her emotions.                     & It's okay to cry.                                      \\
Blameless (BLA)        & This statement reminds the child that the stressful situation is not his / her responsibility. & It’s not your fault.                            \\
Consolation (CON)      & This statement encourages the child to focus on positive aspects of the situation.             & At least that is over with, for the time being. \\
Reassurance (REA)      & This statement reassurance the child that things will get better.                              & It's going to be alright.     \\ \hline                 
\end{tabular}
\caption{ES statement categories, adapted from \citet{smith2014development,dennis2013towards,liu2021towards}.} \label{tab:EScategories}
\end{table}

\end{document}